\documentclass[runningheads]{llncs}
\usepackage[T1]{fontenc}
\usepackage{graphicx}
\usepackage{booktabs}
\usepackage[misc]{ifsym}
\usepackage{multirow}
\usepackage{amsmath}
\usepackage{color}
\usepackage{amssymb}
\newcommand{\corr}{(\Letter)}
\usepackage{mwe}
\begin{document}

\title{BSViT: A Burst Spiking Vision Transformer for Expressive and Efficient Visual Representation Learning}

\titlerunning{BSViT: A Burst Spiking Vision Transformer}
\author{Hongxiang Peng\inst{1} \and
Dewei Bai\inst{1}  \and
Hong Qu\inst{1}\corr}
\authorrunning{H. Peng et al.}
% % You may leave out the orcidID information, if you want to.
% % Use \corr to indicate the corresponding author. Note the spacing around the \corr command. Only one author can be the corresponding author.

% %N.B.: comment out the \authorrunning{} command for the double-blind version of your paper submitted for review. Later, if your paper is accepted, use the command for the Camera-Ready Version.
% \authorrunning{A.L. Benjamin et al.}
% % If there is one author, write 'A.L. Benjamin'.
% % If there are two authors, write 'A.L. Benjamin and C.C. Broadus Jr.'
% % If there are more than two authors, '[...] et al.' is used.

\institute{School of Computer Science and Engineering, University of Electronic Science and Technology of China, Chengdu 610054, China \email{\{hxpeng,dwbai\}@std.uestc.edu.cn,\{hongqu\}@uestc.edu.cn}}
\toctitle{BSViT: A Burst Spiking Vision Transformer for Expressive and Efficient Visual Representation Learning}
\tocauthor{Hongxiang Peng, Dewei Bai, Hong Qu}
% \and
% Fictional West Coast University, Long Beach CA 90840, USA \email{ccb@fwcu.fake}
% \and
% Secondary European Affiliation, Tiergartenstr. 17, 69121 Heidelberg, Germany
% \email{lncs@springer.com}}

\maketitle              % typeset the header of the contribution

\begin{abstract}
Spiking Vision Transformers (S-ViTs) offer a promising framework for energy-efficient visual learning. However, existing designs remain limited by two fundamental issues: the restricted information capacity of binary spike coding and the dense token interactions introduced by global self-attention.
To address these challenges, this work proposes BSViT, a burst spiking-driven Vision Transformer featuring a Dual-Channel Burst Spiking Self-Attention (DBSSA) mechanism. DBSSA encodes queries with binary spikes and keys with burst spikes to enhance representational capacity. The value pathway adopts dual excitatory and inhibitory binary channels, enabling signed modulation and richer spike interactions. Importantly, the entire attention operation preserves addition-only computation, ensuring compatibility with energy-efficient neuromorphic hardware.
To further reduce spike activity and incorporate spatial priors, a patch adjacency masking strategy is introduced to restrict attention to local neighborhoods, resulting in structure-aware sparsity and reduced computational overhead. In addition, burst spike coding is systematically integrated across the network to increase spike-level representational capacity beyond conventional binary spiking.
Extensive experiments on both static and event-based vision benchmarks demonstrate that BSViT consistently outperforms existing spiking Transformers in accuracy while maintaining competitive energy efficiency.

\keywords{Spiking neural networks \and Vision transformer\and Burst spiking.}
\end{abstract}

\section{Introduction}

Spiking Neural Networks (SNN), as the third generation of neural network models~\cite{maass1997networks}, are considered a key technology for building the next generation of energy-efficient artificial intelligence owing to their event-driven computational paradigm and high biological interpretability~\cite{roy2019towards}. Unlike traditional Artificial Neural Networks (ANNs), SNNs utilize spiking neurons to generate discrete binary spikes for information transmission, which allows them to replace complex Multiply-Accumulate (MAC) operations with sparse Accumulate~(AC) operations. By implementing SNNs, neuromorphic chips such as TrueNorth~\cite{akopyanTruenorthDesignTool2015}, Loihi~\cite{daviesLoihiNeuromorphicManycore2018}, and Tianjic~\cite{pei2019towards} are becoming ideal candidates for ultra-low-power applications.
 
Vision Transformers (ViTs) have become representative of high-performance models, surpassing traditional Convolutional Neural Networks (CNNs) on numerous computer vision tasks due to its self-attention mechanism~\cite{dosovitskiy2020image,vaswani2017attention}. However, the remarkable performance is inextricably linked to their significant computational and energy costs, especially for high-resolution tasks~\cite{dehghani2023scaling}. Currently, integrating the superior performance of Transformers with the extreme energy efficiency of SNNs to build Spiking Transformers has emerged as a highly attractive yet challenging research direction. Recent works such as Spikformer and its variants~\cite{guo2025spiking,yao2023spike,zhou2023spikingformerspikedrivenresiduallearning,zhou2024spikformerv2joinhigh,zhou2023spikformer} have made progress by developing spiking self-attention mechanisms that replace matrix multiplications with addition to reduce computational complexity and eliminate Softmax.
\begin{figure}[t]
    \centering
    \begin{minipage}[b]{0.45\linewidth}
        \centering
        \includegraphics[width=\linewidth]{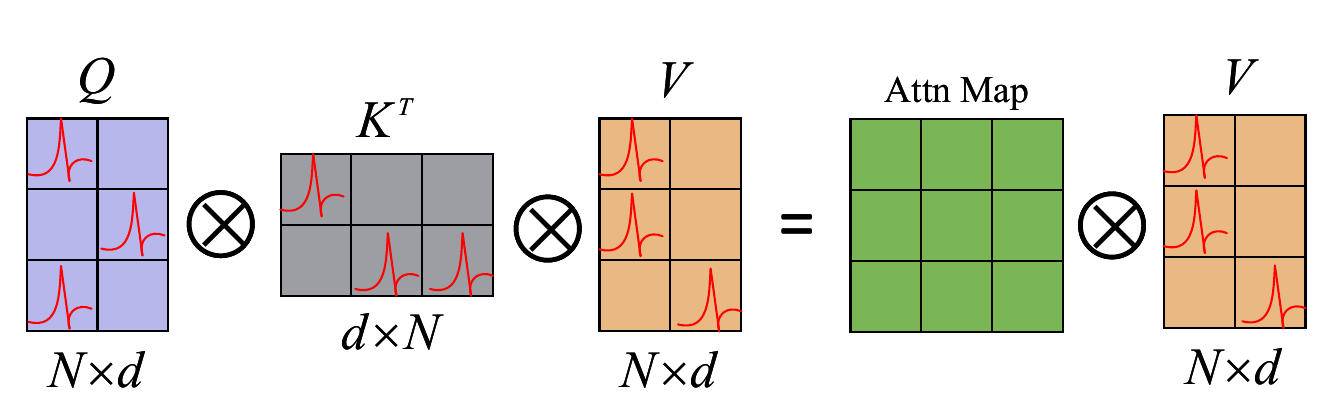}
        \centerline{(a) Spiking Self-Attention (SSA)}
    \end{minipage}
    \hfill
    \begin{minipage}[b]{0.50\linewidth}
        \centering
        \includegraphics[width=\linewidth]{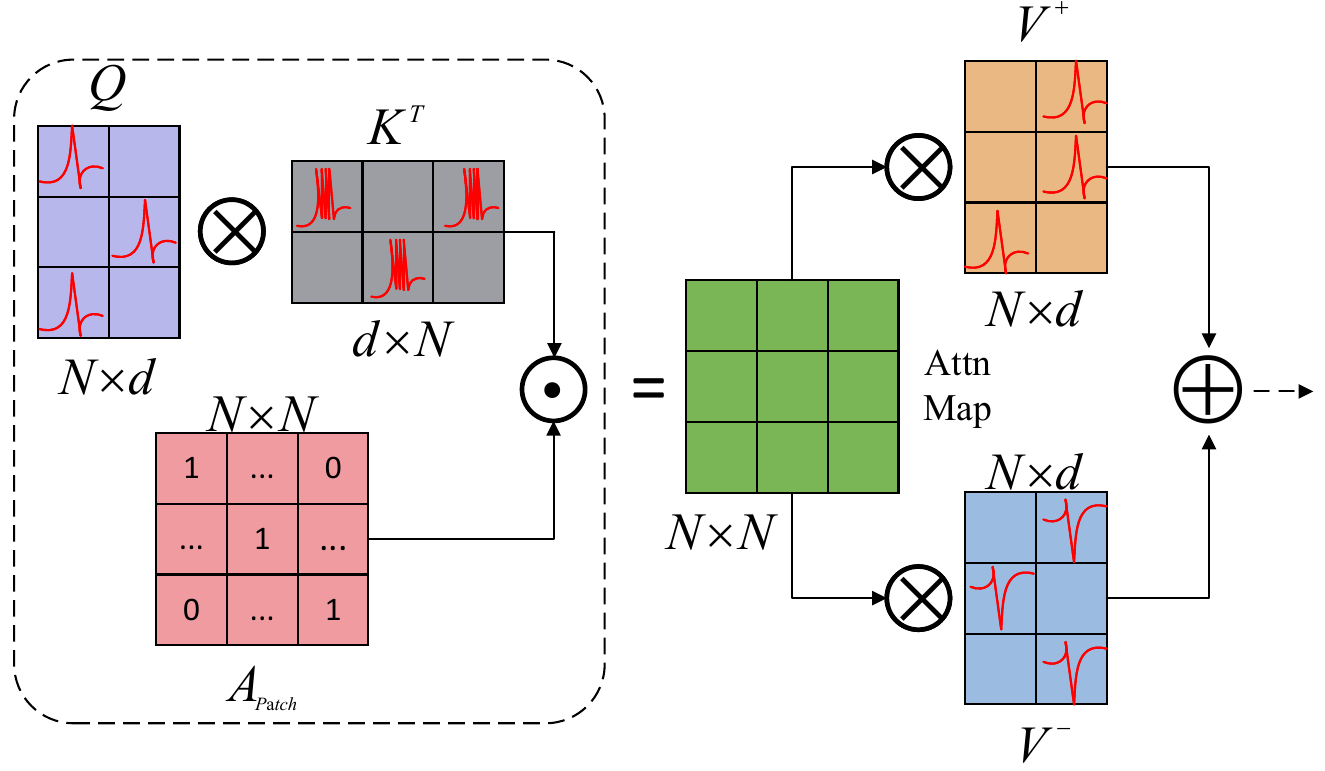}
        \centerline{(b) Dual-channel Burst Spiking Self-Attention}
    \end{minipage}
    \caption{Concept of the Spiking Self-Attention(SSA) and our Dual-channel Burst Spiking Self-Attention(DBSSA). (a) is the vanilla Spiking Self-Attention, only using binary spike matrix to calculate attention map. (b) is our DBSSA mechanism that introduces a burst spiking coded $K$ to increase information capacity and a dual-channel $V$ to capture both excitatory and inhibitory features while preserving addition-only computation. Moreover, the adjacency matrix $A_{patch}$ masks the raw attention map, encouraging local feature aggregation and improving efficiency.}
    \label{fig:1}
\end{figure}
However, these previous methods are faced with a major trade-off between computational efficiency and information capacity. Approaches that strictly use binary spikes achieve the highest efficiency but suffer from severe information loss, while solutions such as incorporating non-spiking activations may compromise the purely event-driven nature of SNNs. To address this issue, we propose the Dual-channel Burst Spiking Self-Attention (DBSSA). As illustrated in Fig. \ref{fig:1}, we keep the binary spike $Q$, utilize a burst spiking neuron to activate $K$, and design a novel dual-channel pattern to generate binary spike $V^+$ and $V^-$ to replace the original $V$. Our design maintains the addition-only operations while improving information capacity by allowing multi-level and signed spiking. Additionally, we employ a patch adjacency matrix to mask the raw attention map, which further reduces energy cost. Finally, we integrated these designs into a transformer architecture based on the burst spiking pattern to construct our Burst Spiking Vision Transformer model. Our main contributions are as follows:

\begin{itemize}
\item We propose a novel Dual-Channel Burst Spiking Self-Attention (DBSSA) mechanism that employs a multilevel burst-coded key for richer similarity representation and a dual-branch binary value path for excitatory and inhibitory modulation, enabling expressive yet addition-only attention computation.
\item We introduce a masked attention scheme based on patch adjacency, introducing spatial inductive bias and structure-aware sparsity to limit spike interactions within local neighborhoods, thereby reducing redundant activity and computational cost without sacrificing accuracy.
\item We develop BSViT, a burst spiking-driven Transformer that systematically integrates burst spiking coding into both feature extraction and attention computation, substantially improving representational power over existing spiking Transformers.
\item Extensive experiments on both static and event-based vision benchmarks demonstrate that BSViT outperforms or matches state-of-the-art spiking Transformers in accuracy, while achieving lower inference latency and comparable computational cost.
\end{itemize}

\section{Related Work}

\subsection{Learning Methods of Spiking Neural Networks}
Two mainstream training paradigms have emerged for Spiking Neural Networks (SNNs). The first is ANN-to-SNN conversion~\cite{buoptimal,dengoptimal,liFreeLunchANN2021a,sengupta2019going}, which maps the firing rate of spiking neurons to approximate ReLU activations in pre-trained ANNs. While this method can produce high-performing SNNs, it often ignores intrinsic temporal dynamics, demands large simulation time steps, and imposes strict constraints on the original ANN architecture. The second paradigm is direct training of SNNs using surrogate gradient methods~\cite{lee2020enabling,wu2018spatio}, which addresses the non-differentiability of spike generation functions and enables end-to-end optimization. Compared to conversion-based approaches, direct training can exploit rich temporal dynamics with significantly fewer time steps and lower latency. With the development of technology, many works have been proposed to gradually improve the accuracy of direct training SNNs. SEW-ResNet and MS-ResNet~\cite{fang2021deep,hu2024advancing,zhengGoingDeeperDirectlytrained2021} successfully improve the shortcut connection to train deep spiking residual networks. Works like TET and Differentiable Spike~\cite{deng2022temporalefficienttrainingspiking,li2021differentiable} optimize the learning algorithms and surrogate gradient. Other methods, such as Multi-level firing, I-LIF and Ternary Spike~\cite{feng2022multi,guo_ternary_2024,luo2024integer}, improve performance by reconstructing the spiking patterns. Our work adopts this direct training paradigm.

\subsection{Spiking Vision Transformers}

Vision Transformers (ViTs) have demonstrated remarkable success across various computer vision tasks, but the floating-point matrix multiplications and Softmax normalization in its self-attention mechanism are energy-consuming and pose fundamental incompatibilities with the SNN paradigm. To bridge the gap between the powerful modeling capacity of Vision Transformers and the energy efficiency of SNNs, recent studies have explored directly trained Spiking Transformers. Spikformer~\cite{zhou2024spikformerv2joinhigh,zhou2023spikformer} introduced the Spiking Self-Attention (SSA) mechanism, in which the query, key, and value vectors are all binarized into spike trains, allowing attention computation to be performed with addition-only operations and removing the need for the Softmax layer. However, SSA still relies on non-spiking residual connections and suffers from limited representational capacity due to binary spike-based attention maps.

To address the non-binarity issue, Spikingformer~\cite{zhou2023spikingformerspikedrivenresiduallearning} applies spiking neurons after the residual connections, ensuring that all inputs to linear and convolutional layers are binary spikes. Spike-driven Transformer~\cite{yao2023spike} further improves computational efficiency by replacing matrix multiplications with Hadamard products. More recently, A²OS²A~\cite{guo2025spiking} proposes a hybrid scheme with binary spiking neurons for queries, ReLU activations for keys, and ternary spikes for values, enhancing expressiveness at the cost of breaking the fully event-driven paradigm due to non-spiking components. In addition,~\cite{yao2025scaling} explores integer activations in place of binary spikes, introducing a new training and inference pattern but not fundamentally addressing the information bottleneck inherent in direct spike-based training.

Unlike prior works, our method combines burst spiking coding, dual-channel signed modulation, and structure-aware sparsity to enhance representational capacity while strictly preserving addition-only, event-driven computation. The details are presented in the following section.

\section{Methodology}

\subsection{Burst Spiking Coding}

Burst spiking is a biologically inspired coding strategy, rooted in the enhanced information transmission observed in mammalian cortical neurons responding to external stimuli~\cite{connorsIntrinsicFiringPatterns1990,izhikevichBurstsUnitNeural2003}. While the Multi-Level Firing (MLF) method~\cite{feng2022multi} simulates bursting by combining multiple LIF neurons with distinct thresholds within a single computational unit, we instead consolidate burst spiking functionality into a single LIF neuron with multiple thresholds, allowing the membrane potential to accumulate within one unit for more efficient burst firing.

In our approach, the burst level at timestep $t$ is determined by quantizing the membrane potential relative to a fixed threshold interval, as follows:
\begin{equation}
\label{eq:lif_burst}
U[t] = \left(\underbrace{\beta U[t-1]}_{\text{decay}} + \underbrace{W X[t]}_{\text{input}}\right) \times  \underbrace{\left(1 - \alpha S_{\text{burst}}[t-1]\right)}_{\text{reset}},
\end{equation}

\begin{equation}
\label{eq:burst_activation_floor}
S_{\text{burst}}[t] = \operatorname{Min}\left(\left\lfloor \frac{U[t]}{V_{\theta}} \right\rfloor, n\right)
\end{equation}

\begin{equation}
    \label{eq:convert}
    W^{(\ell+1)} S_{\text{burst}}^{(\ell)}[t]=\sum_{k=1}^{{S_{\text{burst}}}^{(\ell)}[t]}
{W}^{(\ell+1)},
\end{equation}
where $V_{\theta}$ denotes the interval between consecutive membrane potential thresholds, and $n$ is the maximum allowed burst level. $S_{\text{burst}}[t] \in {0,1,\dots,n}$ thus encodes the number of spikes emitted at timestep $t$. Conceptually related to the integer spike formulation in I-LIF~\cite{luo2024integer}, we convert integer-valued burst outputs into repeated binary additions at inference, as shown in Eq.~\ref{eq:convert}.

This floor-based quantization mechanism offers a compact and hardware-friendly approach for implementing burst firing in spiking neurons. Moreover, inspired by~\cite{fang2021incorporating}, we make the decay rate $\beta$ and the soft reset factor $\alpha$ learnable, allowing both the membrane potential dynamics and spike emission patterns to be adaptively tuned for different tasks.

\begin{figure}[t]
    \centering
    \includegraphics[width=1\linewidth]{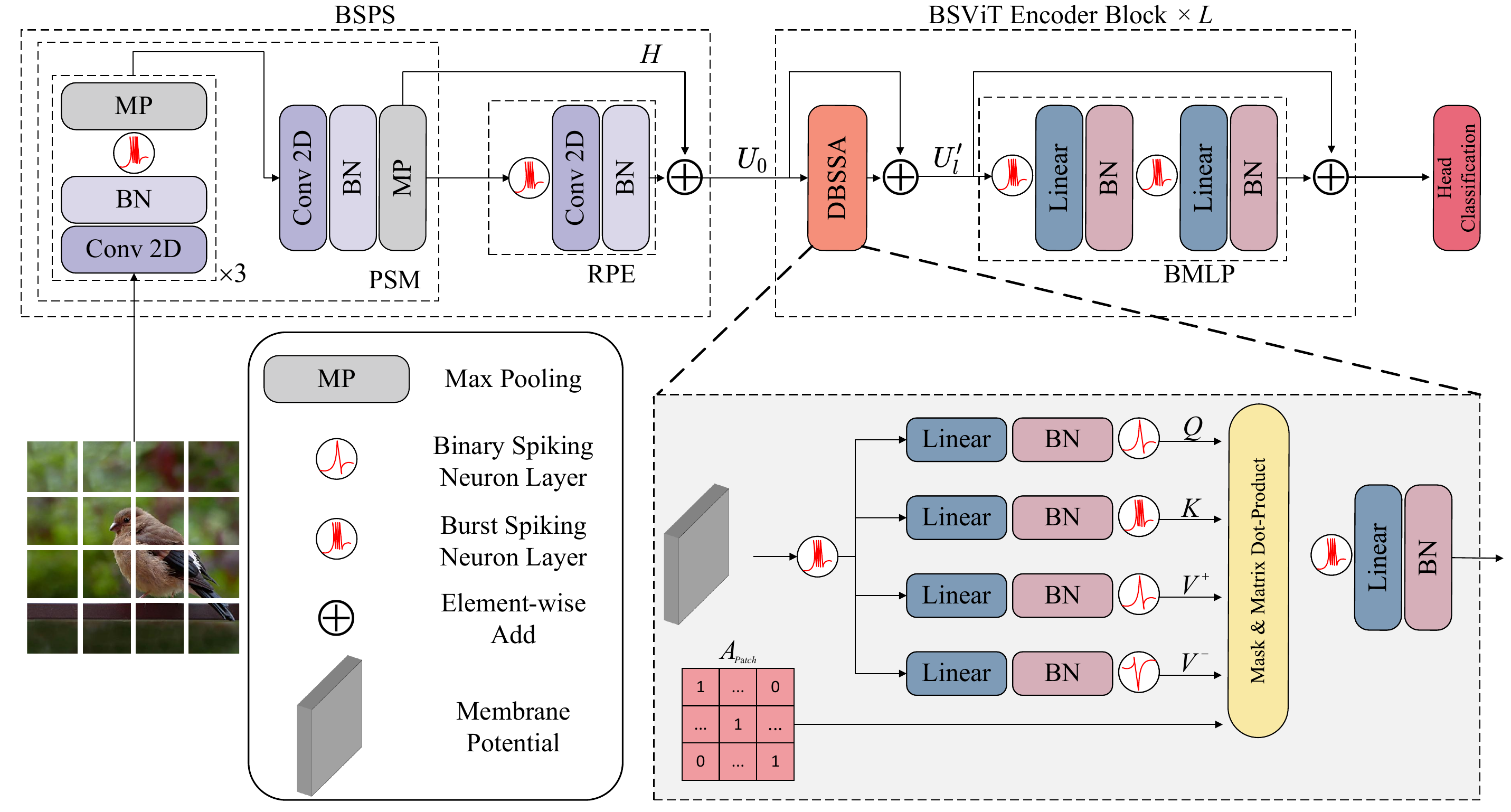}
    \caption{The overview of BSViT.}
    \label{fig:bsvit_architecture}
\end{figure}
\subsection{Overall Architecture}
To effectively integrate the advantages of burst spiking neurons and attention mechanisms into an energy-efficient neural architecture, we follow the design philosophy of Spike-driven Transformer~\cite{yao2023spike}. Our proposed model comprises four key components: Bursting Spiking Patch Splitting~(BSPS), Dual-channel Burst Spiking Self-Attention~(DBSSA), Bursting Multi-Layer Perceptron~(BMLP), and a linear classification head. An overview of the BSViT architecture is illustrated in Fig.~\ref{fig:bsvit_architecture}. Given a 2D image sequence $X \in \mathbb{R}^{T \times C \times H \times W}$(where T, C, H, and W represent time steps, channels, height, and width, respectively), the BSPS module converts it into a 1D sequence of multi-level spiking patches. The Patch Splitting Module~(PSM) first uses a multi-stage convolutional stem to perform initial feature extraction and downsampling on the input image, and next we employ a conditional position embedding generator to generate Relative Position Embeddings (RPE)~\cite{chu2021twins,yao2023spike}. Together, the SPS part is written as:

\begin{equation}
    H=\operatorname{PSM}(X), H \in \mathbb{R}^{T \times N \times D},
\end{equation}

\begin{equation}
    S=\mathcal{SN}_{\text{burst}}(H), S \in \mathbb{R}^{T \times N \times D},
\end{equation}
\begin{equation}
    U_{0}=H+\operatorname{BN}(\operatorname{Conv_{2d}}(S)), U_{0} \in \mathbb{R}^{T \times N \times D},
\end{equation}
where $H$ and $U_0$ are the output membrane potential tensor of PSM and BSPS respectively, and $\mathcal{S N}_{\text{burst}}$ denotes our burst spiking neuron layer. Then we pass the $U_0$ to the \textit{L}-block burst spiking-driven Transformer, which contains DBSSA and BMLP modules. Residual connections are applied in both the DBSSA and BMLP block. The DBSSA block uses binary spike $Q$, burst spike $K$, and dual-channel binary spike represented $V$ to capture more information while keeping addition-only in the core mechanism of attention map calculation. The BMLP uses burst spiking coding to further enhance the information capacity of the attention mechanism. A global average-pooling (GAP) is applied to the processed encoder features and outputs the D-dimension feature which will be sent to the fully-connected-layer classification head (CH) to output the prediction Y. The three parts DBSSA, BMLP and CH can be written as follows:
\begin{equation}
    S_{0}=\mathcal{S N}_{\text{burst}}\left(U_{0}\right), S_{0} \in \mathbb{R}^{T \times N \times D},
\end{equation}
\begin{equation}
    U_{l}^{\prime}=\mathrm{DBSSA}\left(S_{l-1}\right)+U_{l-1}, U_{l}^{\prime} \in \mathbb{R}^{T \times N \times D}, 
\end{equation}
\begin{equation}
    S_{l}^{\prime}=\mathcal{S N}_{\text{burst}}\left(U_{l}^{\prime}\right), S_{l}^{\prime} \in \mathbb{R}^{T \times N \times D},
\end{equation}
\begin{equation}
    S_{l}=\mathcal{S N}_{\text{burst}}\left(\operatorname{BMLP}\left(S_{l}^{\prime}\right)+U_{l}^{\prime}\right), S_{l} \in \mathbb{R}^{T \times N \times D}, 
\end{equation}
\begin{equation}
    Y=\mathrm{CH}\left(\operatorname{GAP}\left(S_{L}\right)\right),
\end{equation}
where $l=1 \ldots L$.

\subsection{Dual-Channel Burst Spiking Self-Attention (DBSSA)}

The Dual-Channel Burst Spiking Self-Attention (DBSSA) mechanism is designed to be entirely spike-based and addition-only, while significantly enhancing the representational capacity. DBSSA leverages binary spikes for $Q$, burst spikes for $K$, and a novel dual-channel binary spike representation for $V$ to preserve both efficiency and expressiveness.

Specifically, $Q$ is encoded using a binary spiking neuron as in prior work. To capture richer similarity patterns, $K$ is encoded via a burst spiking neuron, which inherently supports multilevel spiking. Inspired by the functional segregation of excitatory and inhibitory populations in biological neural circuits~\cite{bittner2017population,wilson1972excitatory}, we design a dual-channel scheme for $V$ by splitting it into excitatory ($V^+$) and inhibitory ($V^-$) components, each generated by an independent binary spiking neuron. This excitatory-inhibitory dynamic acts as a critical filtering mechanism that actively suppresses redundant attention scores, thereby significantly improving the signal-to-noise ratio in the aggregated attention maps. The formulations are as follows:

\begin{equation}
Q = \mathcal{SN}_{\text{binary}} \left( \mathrm{BN}(X W_Q) \right),
\end{equation}
\begin{equation}
K = \mathcal{SN}_{\text{burst}} \left( \mathrm{BN}(X W_K) \right),
\end{equation}
\begin{equation}
V^+ = \mathcal{SN}_{\text{binary}} \left( \mathrm{BN}(X W_{V^+}) \right),
\end{equation}
\begin{equation}
V^- = -\mathcal{SN}_{\text{binary}} \left( \mathrm{BN}(X W_{V^-}) \right),
\end{equation}
where $\mathcal{SN}_{\text{burst}}$ denotes burst spiking encoding with an LIF neuron. The DBSSA output is computed as:

\begin{equation}
\mathrm{DBSSA}(X) = \mathcal{SN}_{\text{burst}}\left( \left[ QK^\top V^+ + QK^\top V^- \right] \cdot s \right),
\end{equation}
where $s$ is a learnable scaling factor replacing the conventional Softmax.

In this formulation, $Q \in \{0,1\}$ allows the attention matrix $QK^\top$ to be computed via addition-only operations, eliminating the need for expensive floating-point multiplications. The burst-coded $K$ not only alleviates the information bottleneck of binary attention maps but also ensures the event-driven, temporal fidelity of spiking computations. As both $Q$ and $K$ are non-negative, the scaling factor $s$ suffices for soft normalization.

In the value path, $V^+$ and $V^-$ enable bipolar feature modulation, allowing both positive and negative contributions in the output, similar to ANNs, without breaking spike-based constraints. Though this dual-channel design increases parameter count, it introduces only minimal additional energy overhead while significantly improving expressivity and preserving spike sparsity. In the actual calculation process, we first multiply $ QK^\top$ by the original $V^-$ result generated by the LIF neurons and then negate it, thus avoiding the extra cost caused by processing the sign bit. Notably, all operations in DBSSA—including attention computation and feature modulation—are expressed as integer additions, making the design biologically plausible and hardware efficient.

\begin{figure}[t]
    \centering
    \includegraphics[width=0.3\linewidth]{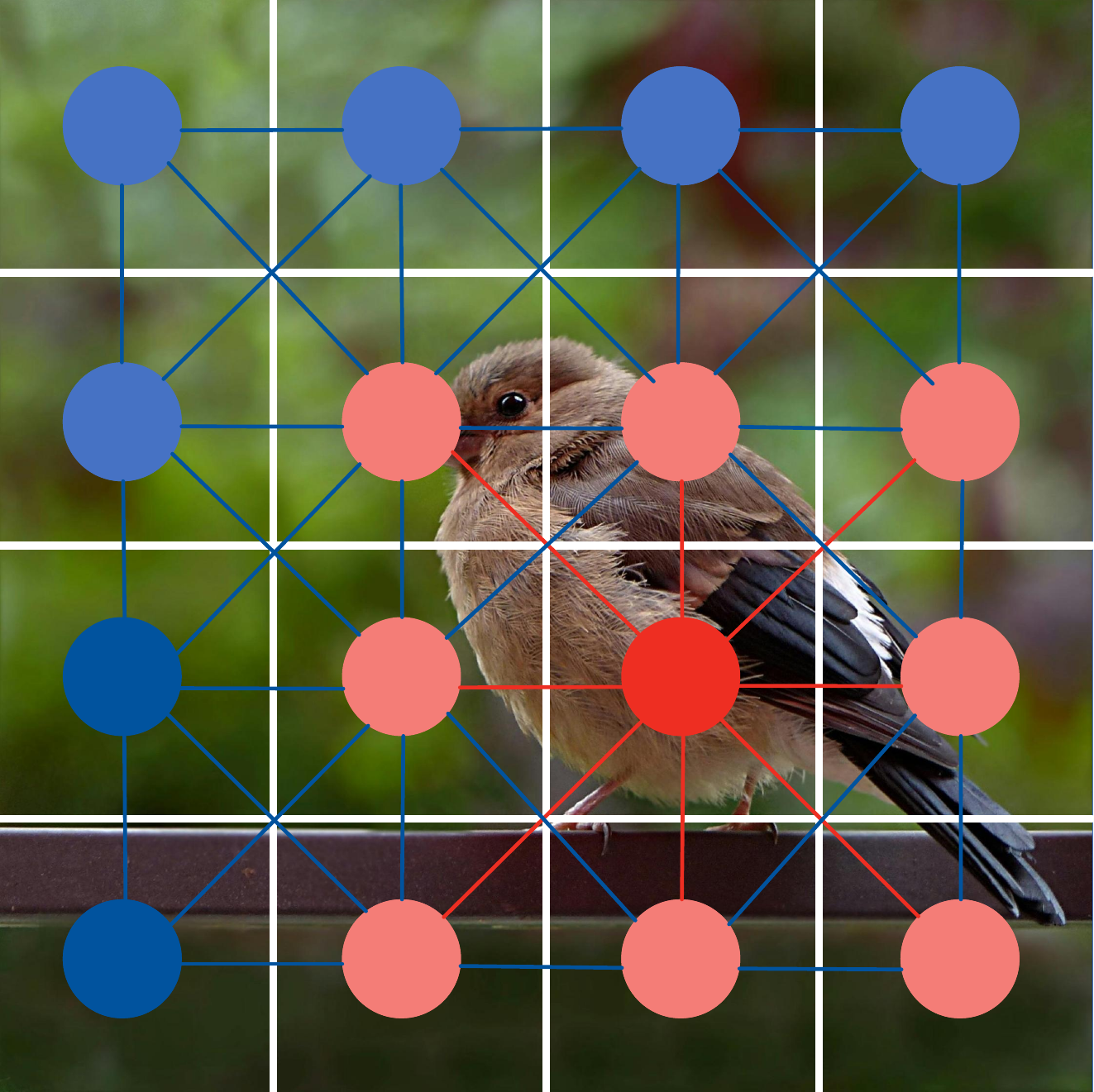}
    \caption{The neighbors of each patch in an image.}
    \label{fig:neighbor}
\end{figure}

% \subsection{Spatial Attention Masking via Patch Adjacency Matrix}
\subsection{Energy-Efficient Attention via Patch Adjacency Masking}

The fully self-attention mechanism interactions between all pairs of patches, resulting in a global receptive field. While powerful, this full attention is computationally expensive and overlooks the strong spatial locality that serves as an inductive bias in vision tasks. To further reducing unnecessary global interactions and promoting localized feature extraction of DBSSA, we introduce a masking strategy that constrains attention computation to local patch neighborhoods.

We first construct a patch adjacency matrix \( A_{\text{Patch}} \in \{0,1\}^{N \times N} \), where \( N = H' \times W' \) denotes the number of patches arranged in a 2D grid of size \( H' \times W' \), as output by the BSPS module. Each entry \( A_{\text{Patch}}[i, j] = 1 \) if patch \( i \) and patch \( j \) are spatially adjacent in the grid, defined by a self-inclusive 8-connected neighborhood shown in Fig.~\ref{fig:neighbor}; otherwise, \( A_{\text{Patch}}[i, j] = 0 \).

This adjacency matrix is then used to mask the raw attention map \( \operatorname{Attn} \) (computed via queries \( Q \) and keys \( K \)) by applying an element-wise (Hadamard) product:
\begin{equation}
    \label{eq:mask}
    \operatorname{Attn}_{\text{masked}} = \operatorname{Attn} \odot A_{\text{Patch}},
\end{equation}
where \( \odot \) denotes the Hadamard product. This masking restricts each patch to attend only to its local neighborhood, encouraging the model to focus on local feature learning in shallow layers—an approach shown to be effective in prior vision studies~\cite{hassani2023neighborhood,kipf2016variationalgraphautoencoders,min2022neighbour}. Additionally, the masked attention map becomes sparser, which improves the model’s energy efficiency, as discussed in Table~\ref{tab:energy_estimation}.

\newcommand{\thead}[1]{\shortstack{#1}}
\begin{table}[t]
\centering
% \small
\caption{Performance comparison between BSViT and existing approaches on CIFAR10/100. Param refers to the number of parameters. BSViT-\textit{L-D} refers to the model with \textit{L} encoder blocks and \textit{D} feature embedding dimensions.}
\label{tab:comparison}
\begin{tabular}{lcccc}
\toprule
% 3. 将长标题换行显示
Methods & \thead{Param\\(M)} & \thead{Time\\Step} & \thead{CIFAR10\\Acc (\%)} & \thead{CIFAR100\\Acc (\%)} \\
\midrule
Hybrid training~\cite{rathi2020enablingdeepspikingneural} & 9.27 & 125 & 92.22 & 67.87 \\
Diet-SNN~\cite{rathi2020dietsnndirectinputencoding} & 0.27 & 10/5 & 92.54 & 64.07 \\
STBP-tdBN~\cite{zhengGoingDeeperDirectlytrained2021} & 12.63 & 4 & 92.92 & 70.86 \\
TET~\cite{deng2022temporalefficienttrainingspiking} & 12.63 & 4 & 94.44 & 74.47 \\
\midrule
% Spikformer-4-256[64] & 4.15 & 4 & 93.94 & 75.96 \\
% Spikformer-2-384[64] & 5.76 & 4 & 94.80 & 76.95 \\
Spikformer-4-384~\cite{zhou2023spikformer} & 9.32 & 4 & 95.19 & 77.86 \\
% Spikingformer-4-256[63] & 4.15 & 4 & 94.77 & 77.43 \\
% Spikingformer-2-384[63] & 5.76 & 4 & 95.22 & 78.34 \\
Spikingformer-4-384~\cite{zhou2023spikingformerspikedrivenresiduallearning} & 9.32 & 4 & 95.61 & 79.09 \\
SDT-2-512~\cite{yao2023spike}  & 10.28 & 4 & 95.60 & 78.40 \\
Spiking Transformer(A²OS²A)-4-384~\cite{guo2025spiking} & 9.32 & 4 & 96.32 & 79.69 \\
STAtten+SDT-2-512~\cite{lee2025spiking}  & 10.28 & 4 & 96.03 & 79.85 \\
\midrule
\textbf{BSViT-4-384(no-masked)} & 9.91 & \textbf{2} & \textbf{96.55} & \textbf{80.27} \\
\textbf{BSViT-4-384} & 9.91 & \textbf{2} & \textbf{96.25} & \textbf{80.32} \\
\bottomrule
\end{tabular}
\end{table}

\section{Experiment}
We conducted our experiments on static datasets, including CIFAR10/100 \cite{krizhevsky2009cifar10} and ImageNet~\cite{93443fe0aca44aa4b4f773517d12ee30}, and the event-based dataset CIFAR10DVS~\cite{li_cifar10-dvs_2017}. For data augmentation techniques like RandAugment~\cite{cubuk2020randaugment} and random erasing~\cite{zhong2020random} are the same as Spikformer~\cite{zhou2023spikformer}. The default burst level for all experiments is 20.

\subsection{CIFAR}
As shown in Tab.~\ref{tab:comparison}, we compared our method with the prior SoTA works on CIFAR-10 and CIFAR-100 datasets. On the CIFAR-10 dataset, our no-masked BSViT-4-384 model achieves an accuracy of 96.55\%, surpassing all other compared models, including Spikformer (95.19\%), Spike-driven Transformer (95.60\%), and the recent high-performing Spiking Transformer (A²OS²A) (96.32\%). Our standard BSViT model with the adjacency mask also achieves a highly competitive accuracy of 96.25\%. On the CIFAR-100 dataset, our no-masked model's advantage is even more pronounced by reaching an accuracy of 80.27\%. Furthermore, the new state-of-the-art accuracy of 80.32\% achieved by BSViT-4-384 demonstrates the strong generalization capability of the adjacency masking strategy, especially on more complex datasets. Notably, our model requires only 2 time steps to achieve these leading results, indicating a significantly lower inference latency. Furthermore, our model's parameter count of 9.91M is comparable to that of other Transformer-based SNNs, which confirms that the performance gains stem from the superiority of our architectural design, rather than simply an increase in model scale.

\begin{table}[t]
\centering
% --- 优化技巧 ---
% \small
% \setlength{\tabcolsep}{2pt} % 2. 减小列间距
% --- 优化结束 ---
\caption{Ablation study for internal components.}
% 表格主体
\begin{tabular}{lcccc}
\toprule
% 3. 将长标题换行
Methods & \thead{Param\\(M)} & \thead{Time\\Step} & \thead{CIFAR10\\Acc (\%)}  \\
\midrule
Baseline-2-256 & 2.59 & 4 & 94.39  \\
Baseline-2-256+\textbf{Burst-encoded K} & 2.59 & 4 & \textbf{94.64}  \\
Baseline-2-256+\textbf{Dual-channel V} & 2.70 & 4 & \textbf{94.71}  \\
Baseline-2-256+\textbf{DBSSA} & 2.70 & 4 & \textbf{94.94}  \\
Baseline-2-256+\textbf{Masked DBSSA} & 2.70 & 4 & \textbf{94.81}  \\
\textbf{BSViT-2-256 (no-masked)} & 2.70 & 4 & \textbf{95.80}  \\
\textbf{BSViT-2-256} & 2.70 & 4 & \textbf{95.61}  \\
\bottomrule
\end{tabular}
\label{tab:ablation_detailed} % 用于在正文中引用
\end{table}

% 使用 table 环境使其横跨整个页面的宽度
\begin{table}[t]
\centering
% \small
\caption{Results on ImageNet-1k.}
\begin{tabular}{llccc}
\toprule
Methods & Architecture & Param (M) & Time Step & Acc (\%) \\
\midrule
Hybrid training~\cite{rathi2020enablingdeepspikingneural} & ResNet-34 & 21.79 & 250 & 61.48 \\
Spiking ResNet~\cite{hu2021spiking}& ResNet-50 & 25.56 & 350 & 72.75 \\
STBP-tdBN~\cite{zhengGoingDeeperDirectlytrained2021} & Spiking-ResNet-34 & 21.79 & 6 & 63.72 \\
\midrule
\multirow{2}{*}{TET~\cite{deng2022temporalefficienttrainingspiking}} & Spiking-ResNet-34 & 21.79 & 6 & 64.79 \\
 & SEW-ResNet-34 & 21.79 & 4 & 68.00 \\
 \midrule

\multirow{2}{*}{SEW ResNet~\cite{fang2021deep}} & SEW-ResNet-34 & 21.79 & 4 & 67.04 \\
% & SEW-ResNet-50 & 25.56 & 4 & 67.78 \\
%  & SEW-ResNet-101 & 44.55 & 4 & 68.76 \\
 & SEW-ResNet-152 & 60.19 & 4 & 69.26 \\
  \midrule
\multirow{2}{*}{Spikformer~\cite{zhou2023spikformer}} & Spikformer-8-384 & 16.81 & 4 & 70.24 \\
 % & Spikformer-6-512 & 23.37 & 4 & 72.46 \\
 % & Spikformer-8-512 & 29.68 & 4 & 73.38 \\
 & Spikformer-10-512 & 36.01 & 4 & 73.68 \\
 % & Spikformer-8-768 & 66.34 & 4 & 74.81 \\
\midrule
\multirow{2}{*}{Spikingformer~\cite{zhou2023spikingformerspikedrivenresiduallearning}} & Spikingformer-8-384 & 16.81 & 4 & 72.45 \\
 % & Spikingformer-8-512 & 29.68 & 4 & 74.79 \\
 & Spikingformer-8-768 & 66.34 & 4 & 75.85 \\
\midrule
\multirow{2}{*}{SDT~\cite{yao2023spike}} 
% & Spike-driven Transformer-8-384 & 16.81 & 4 & 72.28 \\
 % & Spike-driven Transformer-6-512 & 23.37 & 4 & 74.11 \\
  & SDT-8-512 & 29.68 & 1 & 71.68 \\
 & SDT-10-512 & 36.01 & 4 & 74.66 \\
 % & Spike-driven Transformer-8-768 & 66.34 & 4 & 77.07 \\
\midrule
\multirow{2}{*}{A²OS²A~\cite{guo2025spiking}} & Spiking Transformer-8-384 & 16.81 & 4 & 74.04 \\
 % & Spiking Transformer-6-512 & 23.37 & 4 & 76.22 \\
 % & Spiking Transformer-8-512 & 29.68 & 4 & 76.28 \\
 & Spiking Transformer-10-512 & 36.01 & 4 & 78.66 \\
 \midrule
 \multirow{2}{*}{STAtten~\cite{lee2025spiking}} & SDT-8-512 & 29.68 & 4 & 76.18 \\
 & SDT-8-768 & 66.34 & 4 & 78.11 \\
 \midrule
\multirow{2}{*}{BSViT} 
 & BSViT-8-384 & 18.01 & 1 & \textbf{74.37} \\
 % & BSViT-6-512 & 24.96 & 1 & \textbf{77.22} \\
 % & BSViT-8-512 & 29.68 & 4 & \textbf{76.28} \\
 & BSViT-10-512 & 38.65 & 1 & \textbf{78.98} \\
\bottomrule
\end{tabular}

\label{tab:imagenet_comparison}
\end{table}
\subsection{Ablation Study}
We first verify the individual effectiveness of the proposed components. We start with the baseline from Spike-driven Transformer-2-256~\cite{yao2023spike} and progressively add our designs. The detailed results are shown in Tab.~\ref{tab:ablation_detailed}. 

For the baseline model, the accuracy is 94.39\%. By individually introducing the Burst-encoded $K$ and the Dual-channel $V$ mechanism, the accuracy improves to 94.64\% and 94.71\%, respectively. When both mechanisms are combined to form the full Dual-Channel Burst Spiking Self-Attention (DBSSA), the accuracy reaches 94.94\%, achieving a solid gain of 0.55\% over the baseline. Subsequently, applying the Adjacency Matrix Mask (Masked DBSSA) yields an accuracy of 94.81\%; while marginally lower than the no-masked DBSSA, it introduces structure-aware sparsity that offers significant advantages in computational efficiency and energy cost. Finally, by integrating the improved Bursting Spiking Patch Splitting (BSPS) module and Bursting Multi-Layer Perceptron (BMLP) to form our complete BSViT architecture, the accuracy surges to 95.80\% (no-masked) and 95.61\% (masked), resulting in notable gains of 1.41\% and 1.22\% over the baseline, respectively.

\subsection{ImageNet}
Tab.~\ref{tab:imagenet_comparison} presents the comparison with SoTA works on the more challenging ImageNet-1K dataset. Our BSViT model demonstrates exceptional performance across various model scales. Notably, our BSViT-10-512 model achieves a Top-1 accuracy of 78.98\%, representing a significant improvement of 0.32\% over the Spiking Transformer(A²OS²A)-10-512 and surpasses other previous SNN-based Transformer-8-768 models with fewer parameters, which confirms that the performance gains stem from our advanced architectural design rather than a simple increase in model scale. Furthermore, the performance is achieved with an unprecedentedly low latency. When using just a single time step, even our smallest BSViT-8-384 outperforms the  Spike-driven Transformer-8-512 by 2.69 percentage points, showing its potential for real-world, low-latency applications.

\begin{table}[t]
\centering
% \small
\caption{Results on event-based dataset CIFAR10-DVS.}
\begin{tabular}{l cc cc}
\toprule
% --- 第一级表头 ---
% 使用 \multirow 使 "Method" 单元格在垂直方向上跨越两行
Methods  & \thead{Time Step} & \thead{Acc (\%)}  \\

\midrule
% LIAF-Net [58]\textsuperscript{TNNLS-2021} & 10 & 70.4 & 60 & 97.6 \\
% TA-SNN [59]\textsuperscript{ICCV-2021} & 10 & 72.0 & 60 & 98.6 \\
% Rollout [60]\textsuperscript{Front. Neurosci-2020} & 48 & 66.8 & 240 & 97.2 \\
% DECOLLE [61]\textsuperscript{Front. Neurosci-2020} & - & - & 500 & 95.5 \\
tdBN~\cite{zhengGoingDeeperDirectlytrained2021} & 10 & 67.8  \\
PLIF~\cite{fang2021incorporatinglearnablemembranetime}& 20 & 74.8  \\
SEW ResNet~\cite{fang2021deep}  & 16 & 74.4  \\
% Dspike [63]\textsuperscript{NeurIPS-2021} & 10 & 75.4 & - & - \\
% SALT [64]\textsuperscript{Neural Netw-2021} & 20 & 67.1 & - & - \\
DSR~\cite{meng2022training}  & 10 & 77.3 \\
% MS-ResNet [7] & - & 75.6 & - & - \\
\midrule
Spikformer~\cite{zhou2023spikformer} & 16 & 80.6  \\
\midrule
Spikingformer~\cite{zhou2023spikingformerspikedrivenresiduallearning} & 16 & 81.3  \\
\midrule
Spike-driven Transformer~\cite{yao2023spike} & 16 & 80.0 \\
\midrule
BSViT(no-masked) & 16 & 81.9 \\
BSViT & 16 & 81.5 \\
\bottomrule
\end{tabular}

\label{tab:neuromorphic_results}
\end{table}

\subsection{Event-based Datasets}
CIFAR10-DVS is a neuromorphic dataset derived from CIFAR-10 by recording CIFAR-10 images with a Dynamic Vision Sensor (DVS) camera. We use data augmentation as described in~\cite{li2022neuromorphic} and the results are summarized in Tab.~\ref{tab:neuromorphic_results}. Among the evaluated methods, our proposed BSViT achieves an accuracy of 81.5\% with 16 time steps, while the no-masked version achieves 81.9\%, both outperforming previous works. The superior performance of BSViT underscores its effectiveness for event-based data processing tasks.

\section{Discussion}

\begin{table}[t]
\centering
% --- 优化技巧 ---
% \small
% --- 优化结束 ---
\caption{Sensitivity analysis on burst levels.}
% 表格主体
\begin{tabular}{lcccc}
\toprule
% 3. 将长标题换行
Burst Level& \thead{Time Step} & \thead{CIFAR10 Acc (\%)}  \\
\midrule
1& 4 & 94.71  \\
2 & 4 & 95.18  \\
5& 4 & 95.52  \\
10& 4 & 95.71  \\
\textbf{20}& 4 & \textbf{95.80}  \\
40 & 4 & 95.76  \\
\bottomrule
\end{tabular}
\label{tab:ablation_level} % 用于在正文中引用
\end{table}

\subsection{Sensitivity Analysis on Burst Level}
To further investigate the impact of multi-level spike coding, we evaluate the performance across different maximum burst levels ($n$) using the no-masked BSViT-2-256 model. As shown in Tab.~\ref{tab:ablation_level}, the configuration restricted to standard binary spikes ($n=1$) achieves a baseline accuracy of 94.71\%. As the burst level increases, the information capacity of the spiking neurons expands, driving a continuous improvement in performance up to $n=20$, where the model reaches its optimal accuracy of 95.80\%. When $n$ continues to increase beyond 20, the accuracy tends to plateau with only marginal fluctuations, and even a slight performance degradation (95.76\%) is observed at $n=40$. This indicates that an excessively large burst capacity does not bring further representational benefits and may instead introduce quantization noise or potential overfitting. Therefore, we adopt $n=20$ as the optimal default setting for our models.

\begin{table}[t]
\centering
\caption{Energy estimation.}
\begin{tabular}{lcccc}
\toprule
Method & Time Step &\#Sops & \#Sign & Energy \\
\midrule
Spikingformer &4 & 178.78M & 0.22M & 14.58uJ \\
\midrule
\multirow{2}{*}{BSViT(no-masked)}  &2 & 255.03M & 0.26M & 20.59uJ \\
&4 & 513.82M & 0.54M & 41.58uJ \\
\midrule
\multirow{2}{*}{BSViT}  &2 & 204.88M & 0.23M & 16.64uJ \\
&4 & 411.43M & 0.47M & 33.44uJ \\
\bottomrule
\end{tabular}

\label{tab:energy_estimation}
\end{table}
\subsection{Energy Estimation}
\label{sub:ee}
We conducted a theoretical energy estimation for both the masked and no-masked BSViT-2-256 models. We choose the Spikingformer-2-256~\cite{zhou2023spikingformerspikedrivenresiduallearning} as the baseline. Since the number of floating point operations of the first rate-encoding layer is the same, to compare, the total energy cost can be calculated as follows:
\begin{equation}
    E_\text{{total}}=N_\text{{SOP}}\times E_\text{{SOP}}+N_\text{{Sign}}\times E_\text{{Sign}},
\end{equation}
where $N_\text{{SOP}}$ and $N_\text{{Sign}}$ represent the number of synaptic operations (SOPs) and the number of spikes respectively, $E_\text{{SOP}}$ and $E_\text{{Sign}}$ are the energy used per SOP and per spike. Following the previous standards~\cite{hu2021spiking,guo_ternary_2024} that $E_\text{{SOP}}$ is 77fJ and $E_\text{{Sign}}$ is 3.7pJ, we calculate the average energy cost per image in CIFAR-10 and put the results in Tab.~\ref{tab:energy_estimation}. Given that BSViT can outperform previous models with fewer time steps, the masked BSViT with 2 time steps incurs only 14.1\% additional energy cost compared with the 4-step Spikingformer baseline, while the no-masked version incurs 41.2\% additional cost for higher accuracy on some datasets.

\subsection{Computational Complexity Analysis}
To objectively evaluate the efficiency of the proposed method, we analyze the computational complexity of our masked DBSSA compared to Vanilla Self-Attention (VSA). Given an input sequence of length $N$ and an embedding dimension $d$, VSA computes global pairwise interactions and aggregates values across the entire sequence. This inherently yields a dense computational complexity of $\mathcal{O}(N^2d)$ operations for both the similarity computation and the value aggregation stages.

In contrast, our DBSSA mechanism optimizes this attention bottleneck in two chronological stages. First, during the raw similarity computation ($QK^\top$), the asymptotic complexity remains $\mathcal{O}(N^2d)$. However, because the query matrix $Q$ consists entirely of binary spikes and the key matrix $K$ utilizes burst spikes, floating-point multiplications are fundamentally replaced by addition-only (Accumulate, AC) operations. Furthermore, considering the dynamic firing rates $R_Q$ and $R_K$ of the query and key spikes respectively, the effective computational cost is significantly reduced to $\mathcal{O}(R_Q R_K N^2d)$ ACs, which is substantially lower than VSA in practice.

Second, a strict spatial inductive bias is enforced via the immediate application of the patch adjacency mask in Eq.~\ref{eq:mask}. This structure-aware sparsity systematically forces all non-local attention interactions to zero. Within an event-driven context, particularly when deployed on neuromorphic hardware, a zero value in the masked attention map explicitly means no spike is transmitted. Consequently, the model completely bypasses the subsequent value aggregation step for those inactive connections. Each patch is only required to execute AC operations for its $k$ non-zero local neighbors rather than all $N$ patches. 

This hardware-aware property allows our model to effectively translate structure-aware sparsity into actual computational savings, strictly bounding the complexity of the value aggregation stage from $\mathcal{O}(N^2d)$ down to $\mathcal{O}(Nkd)$ ACs. Since $k \ll N$, our approach successfully reduces the heaviest phase of the attention computation to a linear complexity, providing a robust theoretical foundation for real-time and energy-efficient processing.

\section{Conclusion}
In this work, we proposed BSViT, a burst spiking-driven Vision Transformer that systematically integrates burst spiking coding into all stages of the vision processing pipeline. Our DBSSA mechanism enhances the representational capacity of spiking attention maps by leveraging burst-encoded keys and a dual excitatory/inhibitory value path for expressive, signed modulation, while maintaining addition-only computation for maximal hardware efficiency. Furthermore, our masked attention scheme, guided by a patch adjacency matrix, injects explicit spatial inductive bias and restricts spike interactions to local neighborhoods, achieving structure-aware sparsity and reducing redundant spike activity. Extensive experiments on both static and event-based benchmarks demonstrate that BSViT achieves higher accuracy than previous spiking Transformers while maintaining competitive energy cost, thereby advancing the scalability and biological plausibility of neuromorphic vision models.

\begin{credits}
\subsubsection{\ackname} 
% This study was funded by the Sichuan Natural Science Foundation (Grant No. 2024NSFSC0520).
This work was supported by the National Natural Science Foundation of China under Grant 62576078.
\subsubsection{\discintname}
The authors have no competing interests to declare that are
relevant to the content of this article.
\end{credits}

\bibliographystyle{splncs04}
\bibliography{mybibliography}

\end{document}